# Threshold Based Indexing of Commercial Shoe Print to Create Reference and Recovery Images

[1]S. Rathinavel, [2]S. Arumugam
[1]Research Scholar, SSCET, Anna University Palani, 625615, India
rathinavel.s@ieee.org
[2]Nandha Engineering College, Anna University, Erode, 638052, India
arumuagamdote@yahoo.co.in

## Abstract

**One of the important evidence in a crime scene that is normally overlooked but very important evidence is shoe print as the criminal is normally unaware of the mask for this. In this paper we use image processing technique to process reference shoe images to make it index-able for a search from the database the shoe print impressions available in the commercial market. This is achieved first by converting the commercially available image through the process of converting them to gray scale then apply image enhancement and restoration techniques and finally do image segmentation to store the segmented parameter as index in the database storage. We use histogram method for image enhancement, inverse filtering for image restoration and threshold method for indexing. We use global threshold as index of the shoe print. The paper describes this method and simulation results are included to validate the method.**

## 1. INTRODUCTION

It is generally understood that marks left by a criminal's footwear at a crime scene may be helpful in the subsequent investigation of the crime. For example, shoeprints can be useful to likely link crime scenes that have been committed by the same person. They can also be used to target the most prolific criminals by allowing officers to be alerted to watch out for particular shoemarks. Also, shoeprints can provide useful intelligence during interviews so that other crimes can be brought to a criminal. Finally, they can provide evidence of a crime if a shoeprint and suspect's shoe match.[1,2]

Sole Mate, Foster & Freeman's database of shoeprints, proved its worth recently during the investigation of a murder of a woman in the kitchen of her Coventry home in the U.K. Officers from West Midlands police station was confirmed by this database that the suspect's shoeprint had been produced by a Mountain Ridge™ shoe, unique to JJB Sports, a UK nationwide shoe retailer, and that there were two models using that particular sole, and this footwear was first available during the summer of 2002. With this information[1,3,4], police officers were able to locate the store in which the footwear was purchased and to trace the shop's copy of the actual receipt issued to the suspect, later confirmed when an empty shoe box was found at the suspect's home.

Unfortunately, when crime scene is improperly secured or is disorganized, the search of the scene often results in this type of impression evidence being overlooked or destroyed. When this type of physical evidence is properly collected and preserved by the crime scene investigator, followed up by a detailed examination by a footwear expert, it can become an important part in proving or disproving a suspect was at the crime scene. In shoe print biometrics the major tasks is the preparation of reference data from the available sources. Normally reference shoe prints of branded commercial shoes are available in the web. But to use these images for feature extraction and recovery from data bases requires certain steps like conversion to gray scale, image enhancement, before storage of reference image and image restoration. A shoe image from a commercial web site is taken and converted it to gray scale to reduce the size of storage then enhanced the image using histogram technique [8]. Section 2 gives the Block diagram. Section 3 deals with conversion from jpeg to bmp and gray scale conversion, section 4 with image enhancement using histogram method, section 5 with image restoration using Wiener filter and constrained least square restoration methods. The relevant results are provided to illustrate the working of the algorithm.

The block diagram of the current work is shown in Fig. 1. The raw jpg image is converted to bmp format using image converter plus then gray scale conversion is carried out using MATLAB routines [11]. Then image enhancement and restoration are carried out.

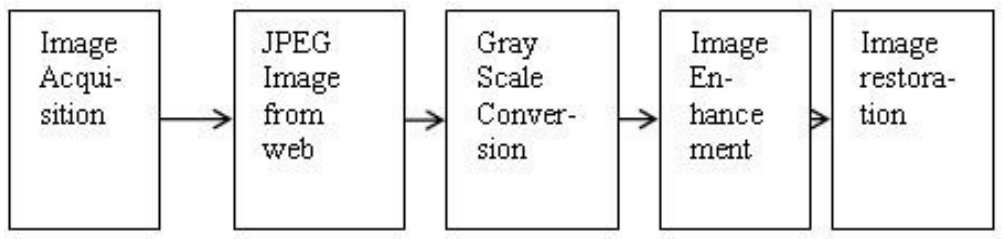

Figure 1 Block diagram depiction

## 2. IMAGE ACQUISITION

Footwear evidence can be found in two forms, impressions and prints [1, 8]. The impression is normally described as three-dimensional impression, such as an impression in mud or a soft material; and the print is described as a print made on a solid surface by dust, powder, or a similar medium.

Footwear evidence, as well as latent fingerprint evidence, is classified into three categories of crime scene prints:

- Visible prints
- Plastic prints
- Latent prints

### 2.1 Visible Prints

A visible print occurs when the footwear steps into a foreign substance and is contaminated by it, and then comes in contact with a clean surface and is pressed onto that surface. This print can be visibly seen by the naked eye without any other aids.

The most common visible prints are prints[1,3] left on a contrasting surface, such as kitchen floor, used. A variety of substances, such as blood, grease, oil or water will leave contrasting prints. This type of print must be photographed, prior to any other methods being used. An electrostatic dust lifter can also be utilized when the evidence is in dust.

### 2.2 Plastic Prints

Plastic prints are impressions that occur when the footwear steps into a soft surface, such as deep mud, snow, wet sand or dirt creating a three-dimensional impression. This type of impressions should be photographed and then cast. These types of impressions are three-dimensional because they allow the examiner to see length, width, and depth [1, 3, 4].

### 2.3 Latent Prints

Latent prints are the most overlooked print and are generally found on smooth surfaces. They can be developed the same way latent fingerprints are. This type of print needs a variety of powders, chemicals and even forensic light sources to make it visible in order to properly be collected. In most cases these prints should also be photographed prior to any recovery process.

However one of the sophisticated methods has been chosen for the investigation and is taken for the comparison with the known set of patterns that are collected from 50 shoes. The known prints are chemical prints obtained by a person stomping on a chemical pad and then a chemical paper which would leave clear print on a paper. All chemical prints are scanned into images at a resolution of standard set of dpi values.

The commercial database gives shoe prints in JPEG/JPG format and to run our program we need it in *.bmp format and hence use image converter plus to convert from jpg to bmp. The image used for the current study is shown in Fig 2

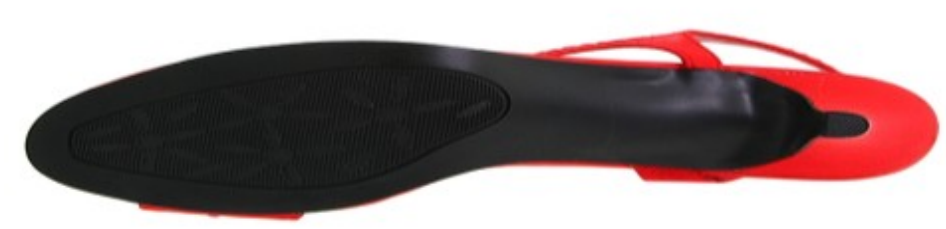

Figure 2 Image considered for study

The gray scale conversion is carried out using the MATLAB routines [8, 10, 11] rgb2ind and ind2gray. The output of such a conversion is shown in Fig. 3

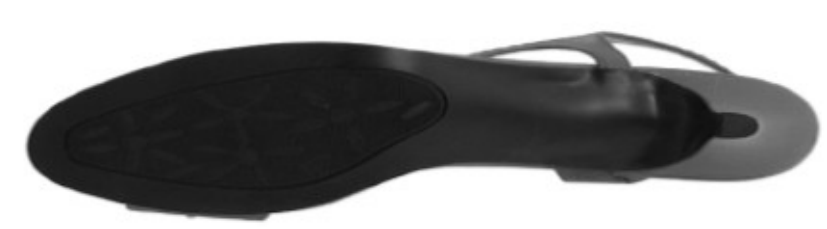

Figure 3 Gray scale image under study

## 3. IMAGE ENHANCEMENT

Since various conversions are done on the image to carry out meaningful feature extraction it is essential we do image enhancement [5,6]. In this study we used the histogram technique to carry out the image enhancement.

The following are the features of the histogram algorithm.

- x0: Converted JPEG input image into *.bmp as a gray level image
- y0: histogram equalized image, each pixel has 8 bits

- L: [0 to L-1] is the number of gray levels. Default L = 256

**Algorithm**

1. Compute the PDF, and then CDF (T(r)) of x0, which is the input image
2. Point-wise gray-level transform according to T(r)

The following MATLAB routines were used hist, round, reshape in this algorithm [2, 7, 9]

A MATLAB program was written and the output of this stage is given in Fig. 4 and Fig. 5

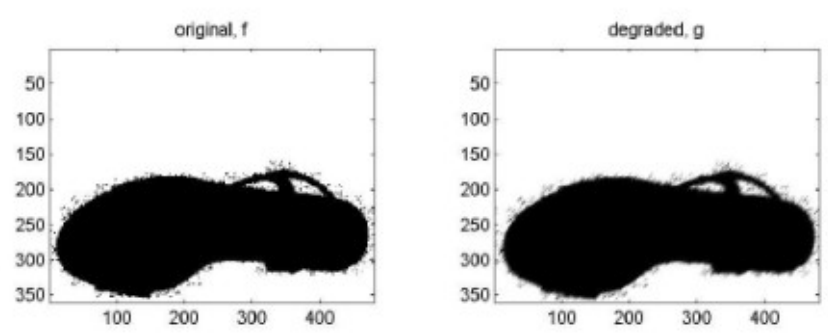

Figure 4 Enhanced image

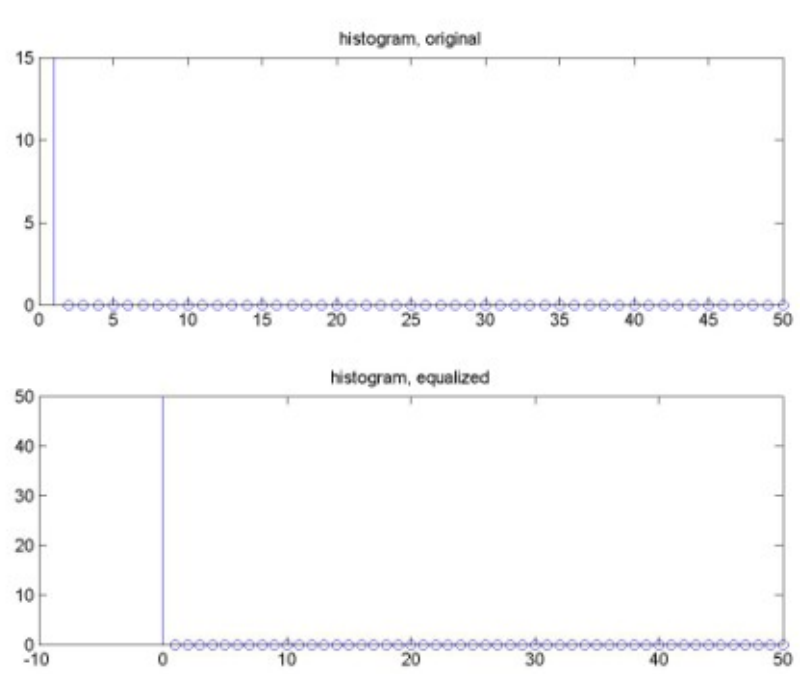

## 4. IMAGE RESTORATION

We carried out the image restoration to understand whether the enhanced signal can be retrieved back from the data base using Wiener and constrained Least Squares[6,8] method. MATLAB code was written and comparisons of the all stages of images are shown in Fig. 6. Motionblur and fft2 are two important MATLAB functions [11] used in this calculation.

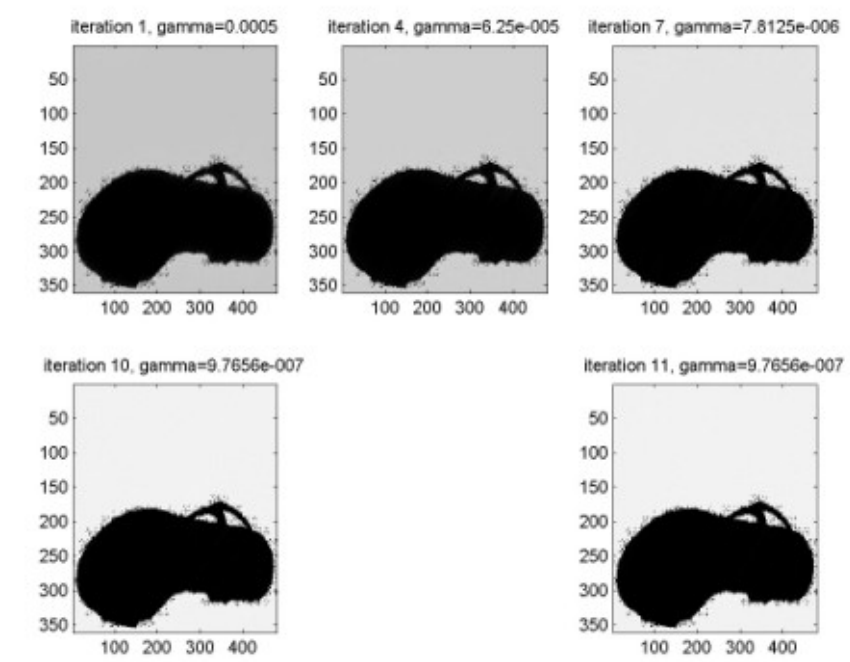

Figure 6 Comparison of all stages

## 5. IMAGE SEGMENTATION

We use threshold method of image segmentation [2, 9] to create the index for a particular reference shoe. The MATLAB simulation was carried out with threshold method of image segmentation with input from restored image and global threshold is obtained which can be used as index for the reference database. The Fig. 7 shows the restored image used as input and Fig. 8 gives the global threshold value that will be used as index of the image database. For the example considered.

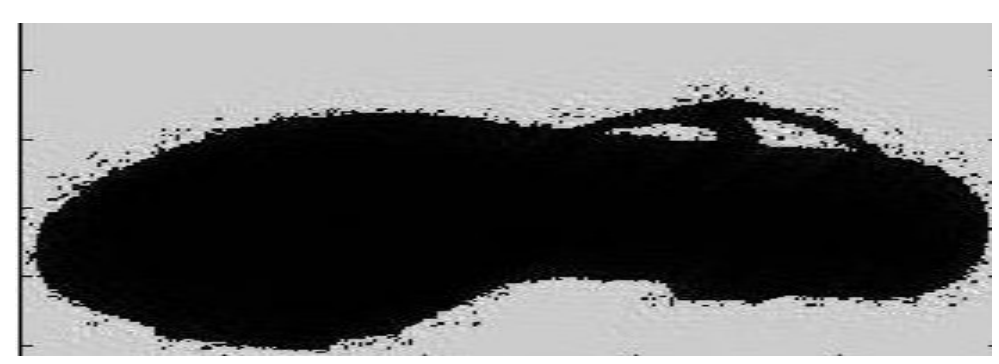

Figure 7 Restored image input for segmentation

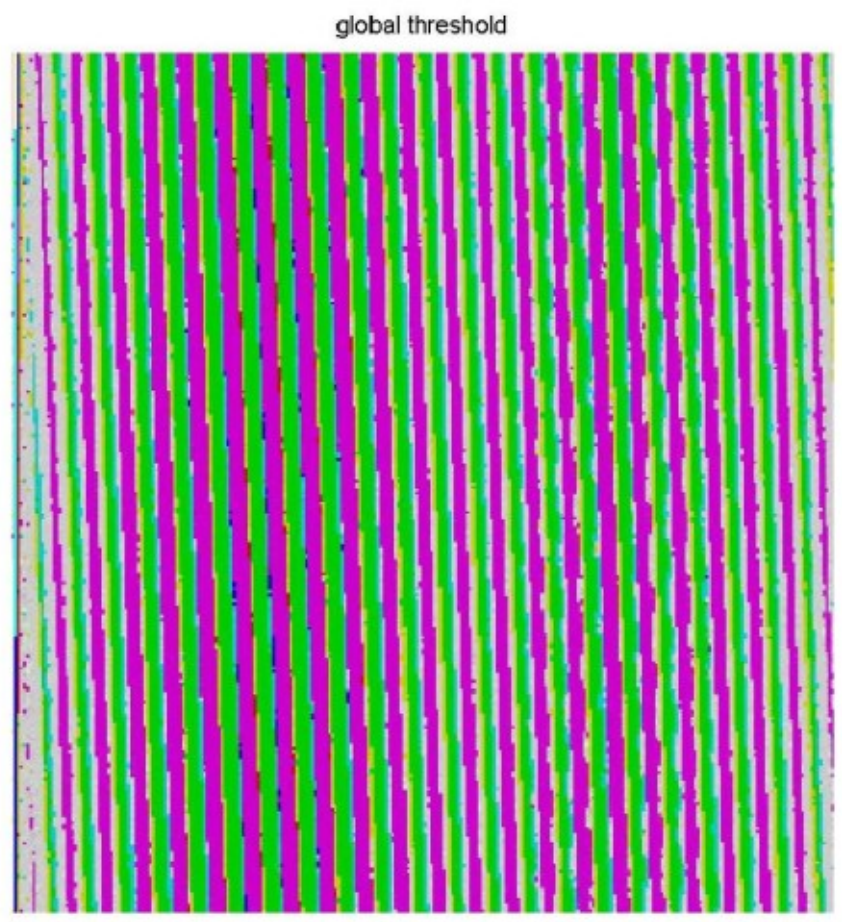

Figure 8 Global threshold of segmented shoe print

## 6. CONCLUSIONS

Thus we have created the indexing global threshold of a commercially available shoe print which is a very important step in the development of reference database for shoe print biometrics. This is part of the on going research and other segmentation methods will be tried to arrive at suitable indexing for the various varieties of shoe commercially available so that the database become richer and able to help in practical situations.

## REFERENCES

[1] W.Bodziac, *Footwear Impression Evidence: Detection, Recovery and Examination,* CRC Press, 2000.

[2] N.Otsu., *A threshold selection method from gray level histogram, IEEE Transaction on Systems, Man and Cybernetics,* 1979.

[3] S. Hildebrand, *Footwear- The Missed Evidence Dwayne*, CLPE Lead Latent Print Examiner, Scottsdale Police crime Lab, no dates or other documentation.

[4] *Imaging For Forensic And Security,* Springer U.S Publisher, ISBN 978-0-387-09531(print)978-0-387-09532-5, www.springerlink.com.

[5] A. Akanu and R. Haddad, *Multi resolution Signal Decomposition,* Academic Press, 1993.

[6] M. Sondhi, *Image Restoration: the removal of spatially-invariant degradation,* Proc. IEEE, vol.6, no. 7, pp.842-853, August 1972.

[7] J. Lee, *Digital Image Enhancement and Noise Filtering by use of Local Statistics,* Proc. IEEE, vol.9, pp.165-168, 1980.

[8] W.K. Pratt, Digital Image Processing, John Wiley and Sons, 1992.

[9] Ernest L. Hall, Computer Image Processing and Recognition, Academic Press, 1979.

[10] TAMAL Bose, Digital Signal and Image Processing, John Wiley & Sons, 2004.

[11] Raefel C.Gonzalez, Richard E. Woods, Steve L. Eddins, "Digital Image Processing using MATLAB",2nd Edition, May 2005.

[12] Alexander, A.; Bouridane, A.; Crookes, D. "Automatic classification and recognition of shoeprints", Image Processing and its Applications, 1999. Seventh International Conference on (Conf. Publ. No. 465), Volume 2, 1999, pp 638-641